\documentclass{article}

\newif\ifneuripsvenue
\newif\ificlrvenue
\newif\ifcameraready %

\neuripsvenuetrue

\usepackage[dblblindworkshop, final]{neurips_2026}

\camerareadytrue

\newcommand{\venuebibstyle}{plainnat}

\usepackage[utf8]{inputenc} %
\usepackage[T1]{fontenc}    %
\usepackage{xurl}           %
\usepackage{hyperref}       %
\usepackage{array}          %
\usepackage{booktabs}       %
\usepackage{multirow}       %
\usepackage{makecell}       %
\usepackage{graphicx}       %
\usepackage{rotating}       %
\usepackage{amsmath}        %
\usepackage{amsfonts}       %
\usepackage{nicefrac}       %
\usepackage{microtype}      %
\usepackage{xcolor}         %
\usepackage{xspace}         %
\usepackage{enumitem}       %
\usepackage{placeins}       %
\usepackage{fvextra}        %
\DefineVerbatimEnvironment{atatprompt}{Verbatim}{frame=single, framesep=2mm,
  fontsize=\small, breaklines, breakanywhere, breaksymbolleft={}}

\usepackage[compact]{titlesec}

\newcommand{\taubench}{$\tau^2$-bench-verified\xspace}
\newcommand{\termbench}{TerminalBench-2\xspace}
\newcommand{\tblite}{TBLite\xspace}
\newcommand{\sigstar}{\rlap{*}}

\newcommand{\company}{\ifcameraready Netflix\else [Anonymized]\fi\xspace}
\newcommand{\companyfn}{\ifcameraready Netflix\else [Anonymized]\footnote{Company identity anonymized only for review; identity will be shared and revealed in any published version.}\fi\xspace}
\newcommand{\codeurl}{\ifcameraready\url{https://github.com/Netflix-Skunkworks/agent-tool-opt-core}\else\url{https://anonymous.4open.science/r/agent-tool-opt-core-2A0D/}\fi\xspace}

\title{AT-AT: Automatic Tool Optimization from Agent Traces}

\workshoptitle{Continual Learning in the Era of Foundation Models and Embodied Agents}

\author{%
  Maya Ravichandran \\ Netflix \And
  Pragya Tripathi \\ Netflix \And
  Dharamendra Kumar \\ Netflix \AND
  Adith Swaminathan \\ Netflix \And
  Nathan Kallus \\ Netflix \AND
  \texttt{\{mravichandran,pragyat,dharamendrak,aswaminathan,nkallus\}@netflix.com}
}

\begin{document}

\maketitle

\begin{abstract}
  Efforts to improve LLM-based agents typically focus on modifying the agent itself, yet success also depends on the environment in which an agent acts. Tools mediate this interaction: a tool's description shapes how an agent calls it, and its implementation determines the results and errors the agent receives afterwards. Many tools, such as APIs and MCP servers, are shared by many agents and maintained by providers who cannot change those agents. We introduce AT-AT (Automatic Tool Optimization from Agent Traces), an agentic tool optimizer that improves a shared toolset offline from historical agent executions while holding the agents fixed. A lightweight coding agent reads agent trajectories and their evaluation results, and selectively edits tool descriptions and, when permitted, implementations. On \(\tau^2\)-bench-verified, AT-AT achieves the highest task completion rates in the Airline and Telecom domains at lower cost than strong prompt- and tool-optimizer baselines, and its Telecom gain is statistically significant under paired permutation tests. Optimizing the tools within a coding agent harness also improves held-out performance on TerminalBench-2 and TBLite. When applied to three production MCP servers at a large enterprise, AT-AT lowered cost or tool error rates on every server and increased held-out task passes on one. By turning accumulated execution traces into updates to shared tool environments, AT-AT provides a path toward continual improvement of agent systems through environment-side learning.
\end{abstract}

\section{Introduction}
\label{sec:intro}

Efforts to improve LLM-based agents typically modify the agent itself: training the underlying model, optimizing its prompts \citep{GEPA,kassianik2026fapo}, or engineering the harness that manages its execution \citep{lee2026metaharness,zhang2026selfharness,weng2026harness}. However, an agent's success is a property not only of the agent, but also of its interaction with the external environment. Tools mediate this interaction, as they define the actions available to the agent and the information returned in response. We study a complementary path to better agent performance: improving the tool environment from accumulated agent experience based on historical execution traces, while holding the agent fixed.

The opportunity for such optimization is large and growing. In an enterprise, an agent's tool environment may include Model Context Protocol (MCP) servers spanning many internal systems. More broadly, nearly any software API can be exposed as an agent-callable tool. Most APIs, however, were designed for human developers or conventional software rather than for the capabilities and limitations of LLM agents. Cloud services can now automatically translate existing APIs into agent-callable tools \citep{microsoft2026expose,wadhwa2025streamline}, but callability does not imply usability by LLMs. An API's agent-facing description, schema, outputs, and errors remain consequential design choices that we aim to automatically optimize. These tools are typically maintained by a provider who cannot change the many agents that call them, but any change the provider makes to a tool reaches all of those agents. This is our primary setting.

A tool has two parts. Its \emph{description}, comprising its name, parameter schema, and usage notes, shapes the agent's expectations \emph{before} a call, while its \emph{implementation} determines the results and errors the agent receives \emph{after} it. Because tool results enter the agent's context, actionable errors can enable recovery and relevant outputs can support better subsequent decisions. We call this mechanism \emph{feedback shaping}: improving the information returned by tools without changing the agent or its evaluation objective. For example, a \taubench Telecom tool that reports a line's data usage and plan limit leaves the agent to work out whether the line has run out of data, while an implementation that also returns the remaining data gives the agent this information directly.

Anthropic reported a 40\% decrease in task completion time after redesigning the tools available to an agent \citep{AnthropicSWEBench}, suggesting that tool design materially affects agent performance. Existing automatic methods primarily optimize the information presented before a tool is invoked: DRAFT~\citep{DRAFT}, Play2Prompt~\citep{Play2Prompt}, EasyTool~\citep{EasyTool}, and ToolObserver~\citep{ToolObserver} use exploration traces or reflection to refine tool descriptions or usage examples. Harness optimizers such as Meta-Harness~\citep{lee2026metaharness} can also edit tool code, but they optimize the harness of a single agent rather than a toolset shared by many agents. In this work, we aim to automatically improve both the descriptions and implementations of shared tools from evaluated agent traces.

We introduce AT-AT (Automatic Tool Optimization from Agent Traces), which converts historical agent experience into selective edits to a toolset. A lightweight coding agent searches evaluated agent traces for recurring failure modes, inspects the tool source, and edits tool descriptions and, when permitted, implementations. A coding agent can search many long traces as needed, and it can read the tool's code to decide whether a failure calls for a description or an implementation fix. AT-AT performs a single offline pass over previously collected executions, requiring no new agent rollouts.

\begin{figure}[t]
    \centering
    \includegraphics[width=\textwidth]{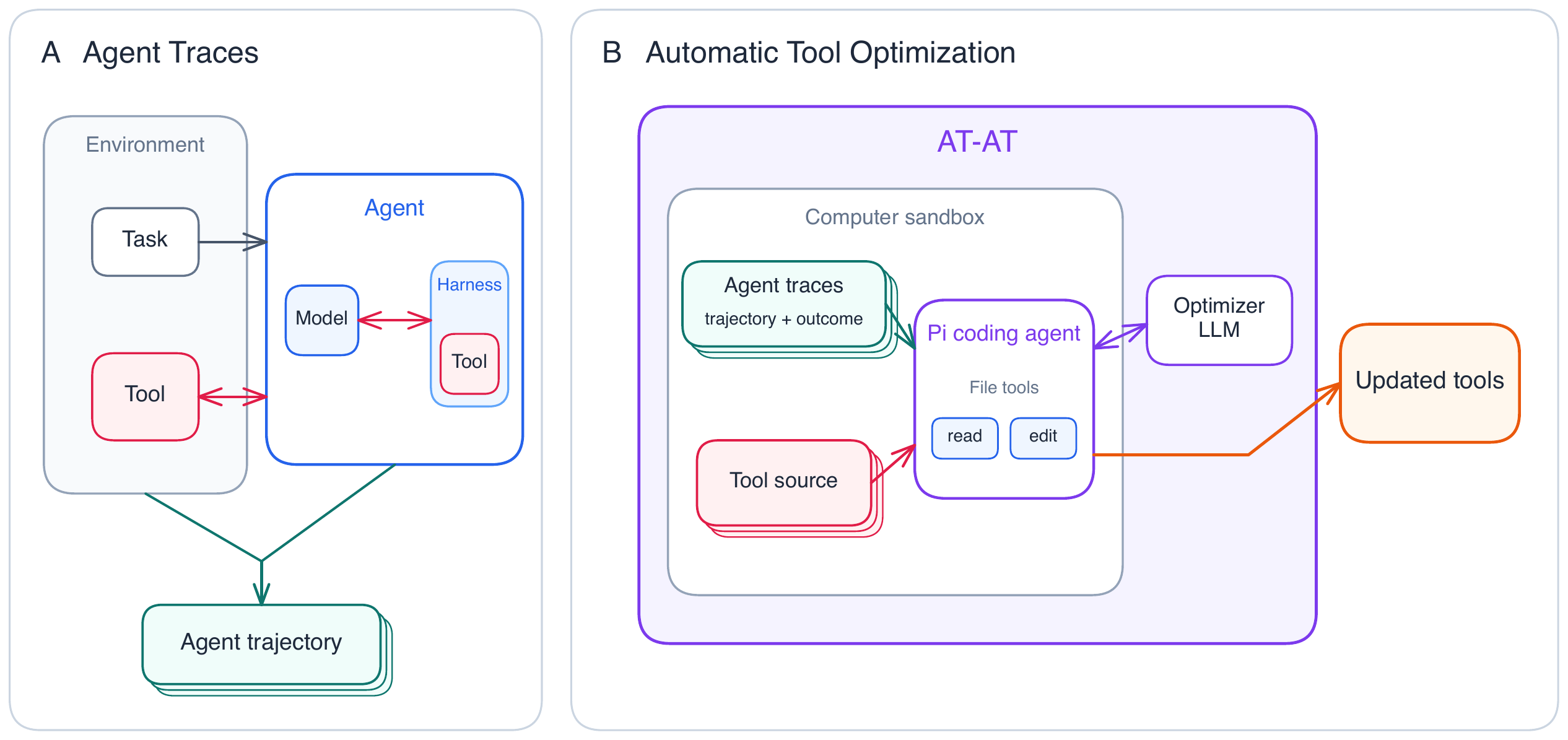}
    \caption{\textbf{AT-AT optimizes tools from historical agent traces.} (A) An agent completes a task by calling tools, implemented in its harness or the external environment, producing a trajectory. (B) In a sandbox holding evaluated traces and the current tool source, a Pi coding agent driven by an optimizer LLM selectively edits tool descriptions and implementations. %
    }
    \label{fig:tool-optimization-overview}
\end{figure}

Trace-driven optimization raises two methodological challenges. First, repeated agent executions can produce materially different outcomes even when the task, model, harness, environment, and decoding configuration are held constant. We evaluate the original and optimized toolsets with 30 trials per held-out task and use paired permutation tests to determine whether observed changes exceed stochastic variation. Second, an optimizer can overfit to a small set of trajectories or tasks, which is a particularly consequential risk for shared tools. We expose AT-AT to traces spanning multiple tasks and repeated trials, direct it toward generalizable edits, and evaluate the resulting toolsets on held-out executions.

We evaluate AT-AT in three settings. On \taubench, AT-AT achieves the highest held-out task completion rate in Airline and Telecom at a lower cost per test simulation than prior prompt and tool optimizers, and its gain of 7.25 percentage points in Telecom is statistically significant. On \termbench and \tblite, optimizing the tools within the OpenCode harness improves held-out pass@1 and pass@3, demonstrating applicability beyond customer service tools. On three production MCP servers at \companyfn, description edits learned from historical logs from production traffic increased held-out task passes on one server and reduced tool errors and cost while maintaining performance on the other two. Together, these results show that the traces used to diagnose agent failures can also be converted into improvements to the environment itself. Because AT-AT can be rerun as traces accumulate, it provides a mechanism for continual improvement, although we evaluate a single update rather than a continual learning loop.

Our contributions are as follows:
\begin{itemize}[topsep=0pt,partopsep=0pt]
    \item We formulate tool optimization for fixed agents from previously collected traces, where a tool's description guides the agent before a call and its implementation shapes the feedback the agent receives after the call (feedback shaping).

    \item We introduce AT-AT, a lightweight coding agent optimizer that translates evaluated historical trajectories into selective edits of tool descriptions and, when permitted, implementations, without new agent executions during optimization.

    \item We evaluate AT-AT on \taubench, \termbench, \tblite, and three production MCP servers, using repeated held-out trials and paired permutation tests on \taubench to separate tool improvements from run-to-run noise.
\end{itemize}
\section{Related Work}
\label{sec:related_work}

\paragraph{Prompt and harness optimization.}
Prompt optimizers search over natural-language or structured prompt programs, using execution outcomes or reflective feedback to guide candidate generation~\citep{SAMMO,GEPA}.
Coding agent optimizers expand this search space: FAPO can modify prompts and multi-step pipeline structure, while Meta-Harness, Self-Harness, and AutoSaddler optimize harness code using execution traces and scores~\citep{kassianik2026fapo,lee2026metaharness,zhang2026selfharness,AutoSaddler}.
Meta-Harness is especially similar in giving an agentic proposer access to an agent's source code, scores, and full traces via a filesystem, but it iteratively searches over the agent harness, which includes the harness's internal tools, rather than a toolset alone~\citep{lee2026metaharness}.
When tools are implemented inside the harness, as in our OpenCode experiments, AT-AT performs a restricted form of harness optimization that edits only the tools. Our primary setting instead updates a shared environmental toolset without changing the agents that consume it, since a tool provider typically does not own those agents.

\paragraph{Manual tool design.}
Anthropic recommends designing and testing tools for LLMs rather than exposing conventional developer-facing APIs
and describes an evaluation-driven workflow in which developers inspect agent transcripts and use a coding agent to revise descriptions and implementations, checking generalization on held-out tasks~\citep{AnthropicSWEBench,Aizawa2025EffectiveTools}.
AT-AT automates such a workflow.

\paragraph{Documentation and schema optimization.}
Most prior methods optimize the prompt-visible surface of a tool.
EasyTool condenses heterogeneous documentation into standardized instructions, while PA-Tool renames schema components to align with a model's pretrained naming conventions~\citep{EasyTool,PATool}.
DRAFT and Play2Prompt interact with tools to refine documentation or synthesize usage examples without labeled tasks~\citep{DRAFT,Play2Prompt}.
A study of 856 tools from 103 MCP servers found at least one quality defect in 97.1\% of tool descriptions, and augmenting the descriptions raised median task success while also increasing execution steps~\citep{hasan2026mcpsmells}.
This still leaves executable tool behavior fixed.

\paragraph{Trace-driven tool optimization.}
Joint context optimization generates training-task rollouts and uses feedback on final answers and tool-call histories to jointly refine agent instructions and tool descriptions, while JTPRO uses rollout-driven reflection to co-optimize global agent instructions and per-tool schemas~\citep{JointToolOpt,JTPRO}.
ToolObserver and DocsChisel repeatedly rerun the agent being optimized and use fixed LLM pipelines to revise only tool documentation from its trajectories.
ToolObserver distills patterns through mini-batching and consensus, whereas DocsChisel uses field-level edits with validation and regression checks~\citep{ToolObserver,DocsChisel}.
AT-AT instead uses a coding-agent optimizer to update shared tools from previously evaluated trajectories without optimization-time agent reruns, and can make contract-preserving implementation edits that change execution-time results, errors, and recovery cues.
ContDa uses self-generated probing traces to continually adapt documentation as tools are added, changed, or deprecated~\citep{ContDa}.
Although AT-AT can be rerun as production logs accumulate, the present work evaluates one update to an existing toolset.

\paragraph{Environment optimization and tool synthesis.}
Other work also improves the environment rather than the agent.
\citet{song2025aegistaxonomyoptimizationsovercoming} derive a taxonomy of agent--environment failures from traces and show that hand-designed environment changes, such as better observability and offloading computation, improve fixed agents; AT-AT instead derives edits automatically from evaluated traces.
Tool-synthesis methods expand the action space by creating new tools from API documentation~\citep{ni2025toolfactory}, composing atomic actions into reusable procedures~\citep{ding2026atomicactionsstandardoperating}, or compiling procedures repeated in execution traces~\citep{Kujanpaa2026ToolMaking}.
AT-AT instead improves tools that are already deployed while preserving the interfaces their callers depend on.
\section{Optimization Method}
\label{sec:optimization_method}

\subsection{Problem Formulation}
\label{sec:problem_setup}
\paragraph{Environmental tool optimization.}
We consider a population of agents \(A \sim \mathcal{A}\). Each agent
\(A=(H,M)\) consists of a fixed harness \(H\) and a fixed large language
model \(M\). The harness includes components such as the system prompt,
code that manages the agent loop and multiple turns, and internal tools
such as file reading and file writing utilities. We also consider a population of tasks \(x \sim \mathcal{X}\), which can be drawn from a variety of domains.

Given an agent \(A\), a task \(x\), and an environmental toolset \(T_E\),
executing the agent generates a trajectory \(\tau\) consisting of the
LLM outputs, user messages, tool invocations, and tool results that
occur during execution. We denote the induced trajectory distribution
by \(p_A(\tau \mid x,T_E)\). Equivalently, for \(A=(H,M)\),
\(p_A(\tau \mid x,T_E)=p_M(\tau \mid H,x,T_E)\). This distribution
captures stochasticity from both the LLM and the environment. The
reward \(r(\tau,x)\) measures task success.

We optimize a shared environmental toolset
\(T_E=(T_{E_D},T_{E_C})\), where \(T_{E_D}\) denotes the tool descriptions
and \(T_{E_C}\) denotes the executable tool code. Environmental tools may
include external functions, APIs, or MCP servers that are not
implemented within the agent harness. Our objective is to select the
toolset that maximizes expected performance across a population of
agents and tasks:

\begin{equation}\textstyle
    \label{eq:tool_opt_obj}
    T_{E_D}^{*}, T_{E_C}^{*}
    =
    \arg\max_{T_{E_D}, T_{E_C}}
    \mathbb{E}_{
        A \sim \mathcal{A},\;
        x \sim \mathcal{X},\;
        \tau \sim p_A(\cdot \mid x,T_{E_D},T_{E_C})
    }
    \left[
        r(\tau,x)
    \right].
\end{equation}

\paragraph{Optimization of tools within a harness.}
For tools implemented within a harness, tool optimization is
a restricted form of harness optimization: only the tool descriptions
and implementations are editable, while all other harness components
remain fixed. Let \(T_H=(T_{H_D},T_{H_C})\) denote the internal tools
being optimized, and let \(H_{-T_H}\) denote the remaining harness
components. The corresponding objective is

\begin{equation}\textstyle
    \label{eq:tool_opt_obj_internal}
    T_{H_D}^{*}, T_{H_C}^{*}
    =
    \arg\max_{T_{H_D}, T_{H_C}}
    \mathbb{E}_{
        x \sim \mathcal{X},\;
        \tau \sim
        p_M(\cdot \mid H_{-T_H},x,T_{H_D},T_{H_C})
    }
    \left[
        r(\tau,x)
    \right].
\end{equation}

We evaluate AT-AT in both tool optimization settings. Our
\taubench{} experiments (Sec.~\ref{sec:tau_bench_expts}) and
production MCP server experiments (Sec.~\ref{sec:mcp_expts})
optimize environmental tools as defined in
Eq.~\ref{eq:tool_opt_obj}. Our coding agent experiments
(Sec.~\ref{sec:coding_agent_expts}) instead optimize tools within
the OpenCode harness.

\subsection{AT-AT Method}
Agent traces on a subset of tasks (either the train split on a benchmark, or historical tasks in a production setting) are constructed. Each agent trace consists of the agent trajectory on a certain task (agent outputs, user messages, tool invocations and results) along with the final evaluation result on the task. The agent traces and tools to edit are provided to the agentic tool optimizer as files in a file system. The optimizer is a lightweight, open-source coding agent, the Pi agent \citep{earendil_pi_coding_agent}. The optimization objective is provided as a system prompt to the Pi agent. An instruction to the optimizer to make generalizable tool edits is also provided in the system prompt. Agent traces are also provided in the user prompt to the optimizer; the first 4,000 characters of each failing run's transcript plus up to three passing runs are included in the first user message to the optimizer LLM. The prompts to AT-AT are provided in Appendix \ref{app:atat-prompts}. The Pi agent makes selective edits to the tool files in the codebase based on what it sees in the agent traces. Once edits are finished, they must pass a validation check. The edited tool files must pass a syntax/compilation/interpreter check, and no files may be added, deleted, or edited outside the editable tool files. If validation fails, the Pi agent's session is resumed with a prompt stating that validation failed, and the agent may revise its edits. The optimizer is given up to five attempts in total to produce a proposal that passes validation. The implementation of AT-AT is available at \codeurl. 
\section{Experiments \& Results}
\label{sec:experiments_and_results}

\subsection{\texorpdfstring{\taubench}{tau2-bench-verified}}
\label{sec:tau_bench_expts}
\(\tau^2\)-bench \citep{barres2025tau2} is a popular benchmark for evaluating LLMs on customer service agent tasks. \taubench \citep{cuadron2025sabersmallactionsbig} is an updated version of \(\tau^2\)-bench in which the tasks were manually reviewed and corrected. In our setup, the user agent and customer service agent LLMs were GPT-5.1 and GPT-4.1, respectively, which is the \taubench default. The default train/test task indices and evaluators from the \taubench repository were used. First, a baseline run of \taubench with the default settings was conducted, with \(n=5\) trials run on each task. The agent trajectories and evaluations were combined to form the baseline agent traces. The same baseline agent traces (train split only) were inputted into all tool optimization methods used, and the tool optimizers were run to generate updated toolsets. The LLM used for all tool optimizers was Claude Opus 4.7. \taubench was rerun using the updated toolsets. Additionally, the average cost of running a task in \taubench was reported, both before and after tool optimizers were applied. This cost was computed using the built-in cost tracker in \taubench. The cost of running the optimizer is also reported and was computed by multiplying the number of tokens used by the cost per token from the LLM APIs. 

\paragraph{Paired permutation testing.} Agent outcomes vary across repeated executions even when the model, harness, tasks, environment, and decoding configuration are held fixed. A difference between two aggregate benchmark scores therefore does not by itself establish that one toolset reliably outperforms another. To account for this evaluation noise, we evaluate the original and optimized toolsets on the same held-out tasks, with 30 trials per task, and compare their outcomes using a paired permutation test. Pairing outcomes within tasks accounts for differences in task difficulty, while repeated trials capture run-to-run variation in agent execution. Under the null hypothesis that the toolset has no effect, the original and optimized condition labels are exchangeable within each pair; permuting these labels produces the null distribution of the difference in task completion rate. We then test whether the observed improvement is larger than would be expected from run-to-run variation. At small sample sizes, exact computation of the paired permutation test \(p\) value is possible, as described in \citet{zmigrod-etal-2022-exact}. However, at our larger sample size of \(n=30\), we estimate the \(p\)-value using Monte Carlo permutation sampling. Each value we compute is the comparison of the optimized toolset result compared to the original, unoptimized toolset.

\paragraph{Baseline optimizers.} We compare to DRAFT \citep{DRAFT}, ToolObserver \citep{ToolObserver}, and GEPA \citep{GEPA}. DRAFT is an online method requiring interaction with live tools, while our \taubench experiment setup provides pre-collected agent traces. Hence, our implementation contained the offline portions of DRAFT, applying DRAFT's Analyzer and Rewriter stages once to the past agent traces. Similarly, we adapt ToolObserver to a fixed offline setting by applying its minibatch reflection and consensus merging once to pre-collected trajectories, omitting iterative exploration and trajectory recollection. We run GEPA \citep{GEPA} by treating the toolset as a prompt to be optimized. GEPA was used with 1 exploration/reflection step to provide a fair comparison to the other methods, which also used 1 exploration/reflection step. 

\paragraph{Overall results.} We report the average reward (pass@1) scores as percentages for each of the three domains: Airline, Retail, and Telecom, both for the baseline (unoptimized) toolset and optimized toolsets produced by AT-AT, our ablations of AT-AT, and the prior baseline optimizers. The results of these experiments can be viewed in Tab.~\ref{tab:tau-bench-full-results}. AT-AT has the highest reward score on the Airline and Telecom domains, and its improvement in Telecom is the only improvement that was statistically significant. In both domains, AT-AT also costs less per test simulation than ToolObserver, the next-best prior method. DRAFT resulted in a statistically significant decrease in performance on Telecom. We contextualize the improvements AT-AT was able to achieve by comparing them to submissions on the official \(\tau^2\)-bench leaderboard \citep{tau2bench_leaderboard}. The 7-point improvement AT-AT achieved on Airline is comparable to the improvement from Claude-3.7-Sonnet (released Feb. 24, 2025; 64\%) to Claude Sonnet 4.5 (released Sep. 29, 2025; 72\%). The 7-point improvement AT-AT achieved on Telecom is comparable to the improvement from Claude-3.7-Sonnet (released Feb. 24, 2025; 49\%) to GPT-5.2 (released Dec. 11, 2025; 57\%). However, the tool edits AT-AT makes can be applied to a variety of agents and potentially improve performance of many agents (discussed in Sec.~\ref{sec:ablations}). 

\paragraph{Costs.} Two types of costs are reported: the cost to run a task in \taubench, and the cost to run the optimizer. Optimization increased the cost to run the benchmark by a statistically significant amount in many cases. We attribute this partially to the increased tool file length, leading to a higher number of tokens that the executor agent needs to process. In addition, variance in costs may be due to prompt cache hit rates. Optimizer costs were similar across optimizers, except for ToolObserver, which incurred substantially higher optimization costs. We attribute this to its minibatch-analysis and consensus-merge structure, which required many more LLM calls than the other optimizers.

\subsubsection{Ablations}
\label{sec:ablations}

\paragraph{Effect of feedback shaping: implementation edits vs description-only edits.} To test our hypothesis that tools' feedback provided to an agent can meaningfully shape its trajectory and improve task success, we ran a version of AT-AT that was constrained to editing tool descriptions only and could not edit tool code. This ablation also omitted agent traces from the user prompt, though they were available in the filesystem. Results are shown in the ``Edit descriptions only'' rows of Tab.~\ref{tab:tau-bench-full-results}. The descriptions-only method performed substantially worse than AT-AT in all domains but Retail (in Retail, all optimized tools performed worse than unoptimized tools, but no difference was statistically significant). A further qualitative analysis comparing tool edits from this ablation is in App.~\ref{app:feedback-shaping}.

\paragraph{Importance of past agent traces.} To test if historical agent traces are a useful learning signal for AT-AT, we ran AT-AT without any historical agent traces available to it. Results are shown in the ``Without past traces'' rows of Tab.~\ref{tab:tau-bench-full-results}. The method without past agent traces performed substantially worse than AT-AT in all domains but Retail (but in the Retail domain, all optimized tools performed worse than unoptimized tools). This evidence supports that past agent traces are a useful learning signal for a tool optimizer. In addition to enabling a tool optimizer to learn about tool failure patterns, historical agent traces may also enable an optimizer to learn about a new domain from data that is automatically gathered, rather than being curated or labeled by humans. 

\paragraph{Importance of coding agent harness.} To investigate the importance of the coding agent setup to the tool optimizer, we ran an experiment with a single-shot LLM-based tool optimizer. Since the LLM-based optimizer did not have the ability to search through a file system itself, the agent traces were only provided in their truncated form in the optimizer user prompt. The optimizer LLM was instructed to output a new tools.py file. Results are shown in the ``Without coding agent harness'' rows of Tab.~\ref{tab:tau-bench-full-results}. The LLM optimizer performed substantially worse than AT-AT in Airline and Telecom.

\paragraph{Generalization across agents.} To investigate our hypothesis that edits to tools made by AT-AT can be useful across a variety of agents and tasks, we generated new baseline transcripts using a variety of customer service agent LLMs: GPT-4.1, GPT-4.1-mini, Inkling, Claude Sonnet 5, and GLM-5.2. AT-AT was run using the train split of these trials. The same 5 LLMs were used when running the test split before and after optimization. This ablation also omitted agent traces from the user prompt, though they were available in the file system. Results are shown in Tab.~\ref{tab:tau-bench-cross-llm}. AT-AT is able to improve reward in Airline and Retail, though the gain is not statistically significant. This is consistent with the hypothesis that AT-AT can improve an environment's overall quality and help a variety of agents.

\begin{table}[t]
  \caption{\textbf{AT-AT achieves the highest test reward in Airline and Telecom.} Held-out results over \(n=30\) trials per task. Test cost is agent plus user-simulator cost per simulation; optimizer cost is the one-time cost of producing the toolset. \(p\): two-sided task-paired permutation test against the domain baseline; \(*\) marks \(p < 0.05\). Bold: best value per domain (optimizer cost excludes baseline).}
  \label{tab:tau-bench-full-results}
  \centering
  \footnotesize
  \setlength{\tabcolsep}{3pt}
  \begin{tabular*}{\textwidth}{@{\extracolsep{\fill}}llrrrrr@{}}
    \toprule
    & Optimizer & \makecell{Test\\reward (\%)} & \makecell{Reward\\\(p\)} & \makecell{Test cost\\(USD/sim.)} & \makecell{Cost\\\(p\)} & \makecell{Optimizer\\cost (USD)} \\
    \midrule
    \multirow{8}{*}{\rotatebox[origin=c]{90}{Airline}}
      & Baseline & 52.17\% & --- & \textbf{0.0618} & --- & --- \\
    \cmidrule(lr){2-7}
      & AT-AT & \textbf{59.17\%} & 0.2305 & 0.0675\sigstar & 0.002155 & 2.7800 \\
      & Without past traces & 50.33\% & 0.3575 & 0.1205\sigstar & 0.000002 & \textbf{0.2330} \\
      & Edit descriptions only & 51.33\% & 0.7609 & 0.1176\sigstar & 0.000002 & 0.3839 \\
      & Without coding agent harness & 52.50\% & 0.9446 & 0.0825\sigstar & 0.000002 & 2.2481 \\
    \cmidrule(lr){2-7}
      & DRAFT & 53.50\% & 0.5988 & 0.0822\sigstar & 0.000002 & 2.4367 \\
      & ToolObserver & 55.83\% & 0.1650 & 0.0870\sigstar & 0.000002 & 16.2171 \\
      & GEPA & 51.50\% & 0.7539 & 0.0748\sigstar & 0.000008 & 0.8042 \\
    \midrule
    \multirow{8}{*}{\rotatebox[origin=c]{90}{Retail}}
      & Baseline & \textbf{77.00\%} & --- & \textbf{0.0707} & --- & --- \\
    \cmidrule(lr){2-7}
      & AT-AT & 74.67\% & 0.2313 & 0.0724 & 0.070629 & 1.9300 \\
      & Without past traces & 75.42\% & 0.3129 & 0.1383\sigstar & 0.000010 & \textbf{0.2174} \\
      & Edit descriptions only & 74.08\% & 0.0992 & 0.1453\sigstar & 0.000010 & 0.5613 \\
      & Without coding agent harness & 74.92\% & 0.1979 & 0.0802\sigstar & 0.000010 & 3.8496 \\
    \cmidrule(lr){2-7}
      & DRAFT & 75.50\% & 0.5296 & 0.0871\sigstar & 0.000010 & 3.9785 \\
      & ToolObserver & 75.92\% & 0.6196 & 0.0926\sigstar & 0.000010 & 42.1400 \\
      & GEPA & 76.08\% & 0.5731 & 0.0798\sigstar & 0.000010 & 0.8765 \\
    \midrule
    \multirow{8}{*}{\rotatebox[origin=c]{90}{Telecom}}
      & Baseline & 29.50\% & --- & 0.0996 & --- & --- \\
    \cmidrule(lr){2-7}
      & AT-AT & \textbf{36.75\%}\sigstar & 0.0033 & \textbf{0.0986} & 0.508165 & 10.6900 \\
      & Without past traces & 30.42\% & 0.6281 & 0.2850\sigstar & 0.000010 & \textbf{0.2888} \\
      & Edit descriptions only & 31.17\% & 0.4547 & 0.2857\sigstar & 0.000010 & 1.1342 \\
      & Without coding agent harness & 30.50\% & 0.6172 & 0.1146\sigstar & 0.000010 & 4.4205 \\
    \cmidrule(lr){2-7}
      & DRAFT & 25.58\%\sigstar & 0.0026 & 0.1259\sigstar & 0.000010 & 4.5630 \\
      & ToolObserver & 31.67\% & 0.4686 & 0.1276\sigstar & 0.000010 & 49.1224 \\
      & GEPA & 29.25\% & 0.9102 & 0.1132\sigstar & 0.000010 & 0.8846 \\
    \bottomrule
  \end{tabular*}
\end{table}

\begin{table}[t]
  \caption{\textbf{AT-AT's tool edits improve held-out reward in 2 of 3 \taubench domains when averaged across five customer service agent LLMs.} Results are averaged across models within each task, then across tasks, with 30 trials per task. Columns and \(*\) are as in Tab.~\ref{tab:tau-bench-full-results}.}
  \label{tab:tau-bench-cross-llm}
  \centering
  \small
  \setlength{\tabcolsep}{3pt}
  \begin{tabular*}{\textwidth}{@{\extracolsep{\fill}}llrrrrr@{}}
    \toprule
    Domain & Optimizer & \makecell{Test\\reward (\%)} & \makecell{Reward\\\(p\)} & \makecell{Test cost\\(USD/sim.)} & \makecell{Cost\\\(p\)} & \makecell{Optimizer\\cost (USD)} \\
    \midrule
    \multirow{2}{*}{Airline}
      & Baseline & 58.00\% & --- & \textbf{0.1135} & --- & --- \\
      & AT-AT & \textbf{59.67\%} & 0.3584 & 0.1239\sigstar & 0.000557 & 2.4029 \\
    \midrule
    \multirow{2}{*}{Retail}
      & Baseline & 73.83\% & --- & \textbf{0.1221} & --- & --- \\
      & AT-AT & \textbf{76.42\%} & 0.1662 & 0.1315\sigstar & 0.000010 & 2.3330 \\
    \midrule
    \multirow{2}{*}{Telecom}
      & Baseline & \textbf{41.42\%} & --- & \textbf{0.2457} & --- & --- \\
      & AT-AT & 40.67\% & 0.7320 & 0.2602\sigstar & 0.000100 & 1.8308 \\
    \bottomrule
  \end{tabular*}
\end{table}

\begin{table}[t]
  \caption{\textbf{AT-AT improves held-out pass@\(k\) on \termbench and \tblite.} Pass@\(k\) from each benchmark's native evaluator before (Baseline) and after (Optimized) optimizing OpenCode's tools, on the train split used for optimization and the held-out test split.}
  \label{tab:termbench-results}
  \centering
  \small
  \setlength{\tabcolsep}{3pt}
  \begin{tabular}{@{}c*{8}{c}@{}}
    \toprule
    & \multicolumn{4}{c}{Train} & \multicolumn{4}{c}{Test} \\
    \cmidrule(lr){2-5} \cmidrule(l){6-9}
    & \multicolumn{2}{c}{\termbench} & \multicolumn{2}{c}{\tblite} & \multicolumn{2}{c}{\termbench} & \multicolumn{2}{c}{\tblite} \\
    \cmidrule(lr){2-3} \cmidrule(lr){4-5} \cmidrule(lr){6-7} \cmidrule(l){8-9}
    \(k\) & Baseline & Optimized & Baseline & Optimized & Baseline & Optimized & Baseline & Optimized \\
    \midrule
    1  & 45.78\% & 48.44\% & 72.80\% & 72.80\% & 49.32\% & \textbf{51.36\%} & 75.60\% & \textbf{77.20\%} \\
    3  & 58.43\% & 58.69\% & 81.38\% & 81.95\% & 59.43\% & \textbf{61.34\%} & 88.65\% & \textbf{89.63\%} \\
    10 & 68.89\% & 68.89\% & 84.00\% & 86.00\% & 68.18\% & \textbf{70.45\%} & 94.00\% & 94.00\% \\
    \bottomrule
  \end{tabular}
\end{table}

\subsection{\termbench \& \tblite}
\label{sec:coding_agent_expts}
To assess whether AT-AT can generalize to a different domain than the customer service agent tasks assessed in \taubench, we ran AT-AT on \termbench and \tblite. \termbench \citep{TerminalBench} is a popular computer use and agentic coding benchmark with long-horizon, multi-turn tasks. \tblite \citep{OpenThoughts-TBLite} is a lightweight alternative to \termbench, designed to track results on \termbench closely while enabling a faster iteration speed and runtime. The toolset optimized was that within OpenCode \citep{anomalyco_opencode}, an open-source coding agent harness. OpenCode contains tools with implementations and descriptions written in TypeScript code, and these tools were the target of the tool optimizer. OpenCode was run on \termbench and \tblite, and the resulting agent traces were split into a train and test set. The train set was then provided to AT-AT to optimize the tools within the OpenCode repository. In these experiments, past traces were omitted from the optimizer user prompt, although they were available in the file system. Finally, the tool-optimized version of OpenCode was rerun on each benchmark to calculate final performance on the train and test splits. The LLM used by AT-AT was Claude Opus 4.7, while the LLM used by the OpenCode agent was Claude Opus 4.5. We report pass@\(k\) as computed by the benchmarks' native evaluators, before and after tool optimization. Results on \termbench and \tblite can be viewed in Tab.~\ref{tab:termbench-results}. On \termbench, AT-AT improves held-out pass@1, pass@3, and pass@10, while on \tblite, AT-AT improves held-out pass@1 and pass@3, while held-out pass@10 is unchanged.

\subsection{Production MCP Servers}
\label{sec:mcp_expts}
\paragraph{MCP and task selection.} We applied AT-AT to production MCP servers at \company.
Three production MCP servers at the company were chosen for optimization due to (1) having a high usage rate, (2) having a sufficient number of unique single-user prompts with actionable failures, and (3) the ability within the company infrastructure to expose and modify the tool descriptions in a sandbox environment. The MCP servers used were the following: \emph{Core Developer Tools} (Dev Tools), which provides broad engineering retrieval across internal documentation, collaboration platforms, and enterprise knowledge sources; \emph{Observability Collaboration Toolbox} (Collab Toolbox), which searches, summarizes, and exports channel and thread content from an enterprise collaboration platform; and \emph{Enterprise Engineering Context and Best Practices} (Eng.\ Context), which provides enterprise-specific engineering guidance from indexed documentation and code. Tasks were defined as an initial agent prompt provided by a user, and tasks were randomly selected from each server's prior two weeks of production traffic. Tab.~\ref{tab:mcp-representative-tasks} summarizes representative tasks.

\paragraph{Experiment procedure.} The list of tasks was randomly partitioned into train/test splits of 10/10, 5/5, and 11/11 tasks for Dev Tools, Collab Toolbox, and Eng. Context and Best Practices, respectively. Across both splits and in the baseline and optimized runs, system prompt, four-user-turn limit, agent model, user simulator, and all other environment settings were held fixed. The agent executing the task used the open-weight DeepSeek V4 Flash model, and the user simulator used Claude Sonnet 4.6. For each task, a baseline run with the original MCP servers was executed, with the task being the initial prompt provided to the executor agent. The baseline run was evaluated as described below. Traces of agent trajectories and evaluation results from the train split of tasks were constructed and provided to AT-AT, which used Claude Opus 4.7 as its optimizer LLM. Past traces were omitted from the optimizer user prompt, although they were available in the file system. After tool optimization, evaluation was repeated on the train and test splits using the same surrounding setup, but now with the optimized tool descriptions.

\paragraph{Evaluation.} Three evaluation metrics were computed for each task. Firstly, a binary pass/fail value for a task was computed via an LLM judge. Secondly, a reward score was computed for each task. \[
r(\tau,x)=
\begin{cases}
1, & \text{if the task-success criteria were met},\\
\frac{1}{2}\frac{n_{\mathrm{accepted}}}{\max(1,n_{\mathrm{eligible}})}, & \text{otherwise},
\end{cases}
\]
Here, \(n_{\mathrm{eligible}}\) is the number of tool calls remaining after harness-blocked and environmental calls are excluded, and \(n_{\mathrm{accepted}}\) is the subset of eligible calls whose result was not an error. This framing provided some partial credit to traces that had fewer tool call failures, while capping unsuccessful traces at a reward of 0.5. 
Thirdly, an LLM judge-based content score was reported to measure answer quality separately from tool execution. An LLM converted each task's success criteria into a fixed rubric: a checklist of 3--7 independently checkable clauses. For \(K\) clauses, the content score was
\[\textstyle
s_{\mathrm{content}}=\frac{1}{K}\sum_{j=1}^{K}q_j,
\qquad q_j\in\{1,0.5,0\}
\]
for clause verdicts met, partially met, and not met, respectively. All evaluation generation and judging used GPT-5.6 Luna. In addition, the total number of tool calls, the number of failed tool calls, the mean cost of agent sessions in each domain, and the cost to run AT-AT were reported. Session cost includes the agent and user simulator but excludes evaluation costs, while optimizer cost is a one-time cost that includes all optimization attempts. Task pairs with a session error either before or after optimization were excluded from efficacy, tool-use, and session-cost metrics. One task in the Eng. Context server was omitted due to a session error and was not used in the optimization or evaluation. 

\paragraph{Results.} Results are shown in Tab.~\ref{tab:offline-mcp-validation-results}. On \emph{Dev Tools}, AT-AT reduced cost, number of tool calls, and number of tool errors, while task completion rate remained unchanged, though content score fell. On \emph{Collab Toolbox}, AT-AT reduced tool calls, tool errors, and execution cost. The mean task reward increased since a larger fraction of eligible tool calls succeeded, but neither arm completed a held-out task, and mean content score declined slightly. On \emph{Eng. Context}, AT-AT improved task completion and content score while reducing tool calls and execution cost. Thus, AT-AT improved efficiency on all three servers and task completion on one. App.~\ref{app:enterprise-mcp-edits} gives paraphrased examples of the resulting tool-description changes and one task's evaluation specification.

\begin{table}[t]
  \caption{\textbf{AT-AT improves task pass rate on one out of three production MCP servers, while reducing tool call error rates and/or cost on all three MCP servers.} Held-out results with original (Orig.) and AT-AT-optimized descriptions; $n$ is the number of scored test tasks. Mean reward and content score are shown as percentages. Session cost covers the agent and user simulator; optimizer cost is one-time.}
  \label{tab:offline-mcp-validation-results}
  \centering
  \small
  \setlength{\tabcolsep}{4pt}
  \begin{tabular}{@{}l*{6}{w{c}{1.4cm}}@{}}
    \toprule
    & \multicolumn{2}{c}{Dev Tools ($n{=}10$)} & \multicolumn{2}{c}{Collab Toolbox ($n{=}5$)} & \multicolumn{2}{c}{Eng.\ Context ($n{=}11$)} \\
    \cmidrule(lr){2-3}\cmidrule(lr){4-5}\cmidrule(lr){6-7}
    Metric & Orig. & AT-AT & Orig. & AT-AT & Orig. & AT-AT \\
    \midrule
    Task pass rate (\%) & 80.0 & 80.0 & 0.0 & 0.0 & 54.5 & \textbf{72.7} \\
    Mean reward (\%) & 90.0 & 90.0 & 22.3 & \textbf{35.1} & 77.3 & \textbf{86.4} \\
    Content score (\%) & \textbf{88.3} & 80.0 & 37.7 & 36.7 & 64.7 & \textbf{69.0} \\
    \midrule
    Tool calls & 176 & \textbf{133} & 121 & \textbf{86} & 114 & \textbf{76} \\
    Tool errors & 15 & \textbf{10} & 73 & \textbf{37} & 0 & 0 \\
    \midrule
    Session cost & \$0.956 & \textbf{\$0.737} & \$0.442 & \textbf{\$0.276} & \$0.288 & \textbf{\$0.235} \\
    Optimizer cost & -- & \$1.34 & -- & \$6.935 & -- & \$7.684 \\
    \bottomrule
  \end{tabular}
\end{table}
\section{Conclusion}
\label{sec:conclusion}

AT-AT optimizes tools to improve agent performance, holding agents fixed and leveraging logged agent traces. This improved performance on multiple benchmarks, including three production MCP servers at \company. Because AT-AT learns from logs that deployed systems already produce, it can be rerun as traces accumulate, storing experience in the environment rather than in the agent.

App.~\ref{app:feedback-shaping} qualitatively analyzes feedback shaping in selected transcripts, and App.~\ref{app:further-analysis} discusses how to handle evaluation stochasticity.

Our results suggest several possible avenues for future work. First, a more explicit model of feedback shaping would be of interest: how changing what a tool returns steers an agent's trajectory. Second, attributing failures to tools, agents, or tasks would bound the headroom for any environment-side intervention and let an optimizer focus on where tools are at fault. Third, a shared tool serves a heterogeneous and evolving population of agents, so tool optimization is fundamentally a problem of generalization across agents and of co-adaptation with agent- and harness-side optimization (App.~\ref{app:harness-opt}), whose interaction is not yet understood. Finally, treating the environment as a site of continual learning calls for guarantees against regression and drift as edits accumulate, and raises the possibility of using tool design to steer agents away from reward hacking and unsafe actions.

\ificlrvenue
\clearpage

\fi

\bibliographystyle{\venuebibstyle}
\bibliography{references}

\appendix
\FloatBarrier
\section{Detailed \texorpdfstring{\taubench}{tau2-bench-verified} Results}
\subsection{Additional Ablations}

\paragraph{Specificity of method to optimizer LLM used.} To assess AT-AT's efficacy with different optimizer LLMs, AT-AT, and ToolObserver for comparison, were run with GPT-5.6 Sol and the open-weight model Kimi K3 with the same \taubench experiment setup. This ablation also omitted agent traces from the user prompt, though they were available in the file system. Results are shown in Tab.~\ref{tab:tau-bench-optimizer-llm-robustness}. AT-AT was able to achieve a statistically significant improvement in Telecom using GPT-5.6 Sol and Kimi K3, although regressions were seen when AT-AT was run with these two models on Airline, though the regressions were not statistically significant. This experiment supports that gains from AT-AT are possible with other models besides Claude Opus 4.7 and that the method is not overfit to a particular optimizer LLM. 

\begin{table}[p]
  \caption{\textbf{AT-AT's improvements are robust to the choice of optimizer LLM in \taubench's Telecom, where both GPT-5.6 Sol and Kimi K3 produce statistically significant gains.} The customer service agent and user simulator are GPT-4.1 and GPT-5.1, respectively; only the optimizer LLM varies. Results are computed over \(n=30\) trials per held-out test task. Test cost is mean agent plus user-simulator inference cost in USD per simulation, and optimizer cost is the one-time cost of producing the candidate toolset. \(*\) marks a statistically significant change from the domain baseline at \(p < 0.05\) under a two-sided task-paired permutation test. Best reward (highest), test cost (lowest), and optimizer cost (lowest non-baseline value) in each domain are bold.}
  \label{tab:tau-bench-optimizer-llm-robustness}
  \centering
  \small
  \setlength{\tabcolsep}{3pt}
  \begin{tabular*}{\textwidth}{@{\extracolsep{\fill}}lllrrrrr@{}}
    \toprule
    Domain & Optimizer & \makecell{Optimizer\\LLM} & \makecell{Test\\reward (\%)} & \makecell{Reward\\\(p\)} & \makecell{Test cost\\(USD/sim.)} & \makecell{Cost\\\(p\)} & \makecell{Optimizer\\cost (USD)} \\
    \midrule
    \multirow{7}{*}{Airline}
      & Baseline & --- & 52.17\% & --- & \textbf{0.0618} & --- & --- \\
    \cmidrule(lr){2-8}
      & \multirow{3}{*}{AT-AT} & Opus 4.7 & \textbf{59.17\%} & 0.2305 & 0.0675\sigstar & 0.002155 & 2.7800 \\
      & & GPT-5.6 Sol & 47.33\% & 0.3743 & 0.1194\sigstar & 0.000002 & \textbf{1.2866} \\
      & & Kimi K3 & 54.33\% & 0.3135 & 0.1246\sigstar & 0.000002 & 4.7419 \\
    \cmidrule(lr){2-8}
      & \multirow{3}{*}{ToolObserver} & Opus 4.7 & 55.83\% & 0.1650 & 0.0870\sigstar & 0.000002 & 16.2171 \\
      & & GPT-5.6 Sol & 52.83\% & 0.7461 & 0.1498\sigstar & 0.000002 & 10.1014 \\
      & & Kimi K3 & 53.33\% & 0.7576 & 0.1619\sigstar & 0.000002 & 8.5355 \\
    \midrule
    \multirow{7}{*}{Retail}
      & Baseline & --- & 77.00\% & --- & \textbf{0.0707} & --- & --- \\
    \cmidrule(lr){2-8}
      & \multirow{3}{*}{AT-AT} & Opus 4.7 & 74.67\% & 0.2313 & 0.0724 & 0.070629 & 1.9300 \\
      & & GPT-5.6 Sol & 74.75\% & 0.2487 & 0.1444\sigstar & 0.000010 & \textbf{0.8445} \\
      & & Kimi K3 & 75.92\% & 0.5848 & 0.1395\sigstar & 0.000010 & 1.4375 \\
    \cmidrule(lr){2-8}
      & \multirow{3}{*}{ToolObserver} & Opus 4.7 & 75.92\% & 0.6196 & 0.0926\sigstar & 0.000010 & 42.1400 \\
      & & GPT-5.6 Sol & 75.83\% & 0.6272 & 0.1490\sigstar & 0.000010 & 27.5875 \\
      & & Kimi K3 & \textbf{77.50\%} & 0.8646 & 0.1774\sigstar & 0.000010 & 23.4634 \\
    \midrule
    \multirow{7}{*}{Telecom}
      & Baseline & --- & 29.50\% & --- & 0.0996 & --- & --- \\
    \cmidrule(lr){2-8}
      & \multirow{3}{*}{AT-AT} & Opus 4.7 & 36.75\%\sigstar & 0.0033 & \textbf{0.0986} & 0.508165 & 10.6900 \\
      & & GPT-5.6 Sol & \textbf{47.58\%}\sigstar & \(<\)0.0001 & 0.2633\sigstar & 0.000010 & \textbf{1.1033} \\
      & & Kimi K3 & 38.42\%\sigstar & 0.0056 & 0.2854\sigstar & 0.000010 & 1.4343 \\
    \cmidrule(lr){2-8}
      & \multirow{3}{*}{ToolObserver} & Opus 4.7 & 31.67\% & 0.4686 & 0.1276\sigstar & 0.000010 & 49.1224 \\
      & & GPT-5.6 Sol & 40.50\%\sigstar & 0.0001 & 0.3083\sigstar & 0.000010 & 32.8944 \\
      & & Kimi K3 & 37.67\%\sigstar & 0.0182 & 0.3219\sigstar & 0.000010 & 25.2612 \\
    \bottomrule
  \end{tabular*}
\end{table}

\paragraph{Importance of generalization prompt and full traces in file system.} To test the contribution of these two additions, we ran the Pi optimizer with only the trace digest in its first user message, without the generalization instruction and without the full per-task traces in its file system. Results are shown in Tab.~\ref{tab:tau-bench-overfitting-fixes}. This variant underperforms AT-AT and does not achieve its gains on Airline and Telecom. Because both components were removed together, this ablation does not separate their individual contributions. 

\begin{table}[p]
  \caption{\textbf{Pi optimizer without the generalization prompt or full file system traces underperforms AT-AT's reward gains in Airline and Telecom.} Held-out results over \(n=30\) trials per task; baseline and AT-AT rows are repeated from Tab.~\ref{tab:tau-bench-full-results} for comparison. Test cost is agent plus user-simulator cost per simulation; optimizer cost is the one-time cost of producing the toolset. \(p\): two-sided task-paired permutation test against the domain baseline; \(*\) marks \(p < 0.05\). Bold: best test reward and test cost per domain.}
  \label{tab:tau-bench-overfitting-fixes}
  \centering
  \footnotesize
  \setlength{\tabcolsep}{3pt}
  \begin{tabular*}{\textwidth}{@{\extracolsep{\fill}}llrrrrr@{}}
    \toprule
    Domain & Optimizer & \makecell{Test\\reward (\%)} & \makecell{Reward\\\(p\)} & \makecell{Test cost\\(USD/sim.)} & \makecell{Cost\\\(p\)} & \makecell{Optimizer\\cost (USD)} \\
    \midrule
    \multirow{3}{*}{Airline}
      & Baseline & 52.17\% & --- & 0.0618 & --- & --- \\
    \cmidrule(lr){2-7}
      & AT-AT & \textbf{59.17\%} & 0.2305 & 0.0675\sigstar & 0.002155 & 2.7800 \\
      & Without generalization prompt or full traces & 51.33\% & 0.8099 & \textbf{0.0605} & 0.363811 & 2.7800 \\
    \midrule
    \multirow{3}{*}{Retail}
      & Baseline & \textbf{77.00\%} & --- & \textbf{0.0707} & --- & --- \\
    \cmidrule(lr){2-7}
      & AT-AT & 74.67\% & 0.2313 & 0.0724 & 0.070629 & 1.9300 \\
      & Without generalization prompt or full traces & 75.75\% & 0.3186 & 0.0710 & 0.683283 & 1.9300 \\
    \midrule
    \multirow{3}{*}{Telecom}
      & Baseline & 29.50\% & --- & 0.0996 & --- & --- \\
    \cmidrule(lr){2-7}
      & AT-AT & \textbf{36.75\%}\sigstar & 0.0033 & \textbf{0.0986} & 0.508165 & 10.6900 \\
      & Without generalization prompt or full traces & 26.92\% & 0.1373 & 0.0993 & 0.750562 & 10.6900 \\
    \bottomrule
  \end{tabular*}
\end{table}

\subsection{Detailed \texorpdfstring{\taubench}{tau2-bench-verified} results}
Tab.~\ref{tab:tau-bench-detailed-results} reports the reward and recurring-cost deltas, 95\% task-cluster bootstrap confidence intervals, and raw paired-permutation \(p\)-values corresponding to the main-paper summary in Tab.~\ref{tab:tau-bench-full-results}.

\begin{sidewaystable}[p]
  \caption{\textbf{Detailed \taubench reward and cost results.} Test reward and recurring inference cost are evaluated over 30 trials per held-out task. Each domain's baseline is reported as a separate row. Deltas are optimized minus baseline and are shown with 95\% task-cluster bootstrap confidence intervals. Reported \(p\)-values are from two-sided task-paired permutation tests; \(*\) marks \(p < 0.05\). Recurring cost is mean agent plus user-simulator cost in USD per test simulation. Optimizer cost is the one-time cost of producing the candidate toolset.}
  \label{tab:tau-bench-detailed-results}
  \centering
  \scriptsize
  \setlength{\tabcolsep}{2.2pt}
  \resizebox{\textheight}{!}{%
  \begin{tabular}{@{}llrrrr@{\hspace{\dimexpr 2\tabcolsep+10pt\relax}}rrr@{}}
    \toprule
    Domain & Optimizer & \makecell{Test\\reward (\%)} & \makecell{Reward delta\\{}[95\% CI]} & \makecell{Reward\\\(p\)} & \makecell{Test cost\\(USD/sim.)} & \makecell{Cost delta\\{}[95\% CI]} & \makecell{Cost\\\(p\)} & \makecell{Optimizer\\cost} \\
    \midrule
    \multirow{8}{*}{Airline}
      & Baseline & 52.17\% & --- & --- & \textbf{0.0618} & --- & --- & --- \\
    \cmidrule(lr){2-9}
      & AT-AT & \textbf{59.17\%} & +0.0700 [-0.0117, +0.1750] & 0.2305 & 0.0675\sigstar & +0.0057 [+0.0025, +0.0095] & 0.002155 & 2.7800 \\
      & Without past traces & 50.33\% & -0.0183 [-0.0533, +0.0133] & 0.3575 & 0.1205\sigstar & +0.0587 [+0.0497, +0.0709] & 0.000002 & \textbf{0.2330} \\
      & Edit descriptions only & 51.33\% & -0.0083 [-0.0483, +0.0350] & 0.7609 & 0.1176\sigstar & +0.0558 [+0.0471, +0.0655] & 0.000002 & 0.3839 \\
      & Without coding agent harness & 52.50\% & +0.0033 [-0.0417, +0.0500] & 0.9446 & 0.0825\sigstar & +0.0207 [+0.0161, +0.0258] & 0.000002 & 2.2481 \\
    \cmidrule(lr){2-9}
      & DRAFT & 53.50\% & +0.0133 [-0.0250, +0.0567] & 0.5988 & 0.0822\sigstar & +0.0204 [+0.0150, +0.0269] & 0.000002 & 2.4367 \\
      & ToolObserver & 55.83\% & +0.0367 [-0.0067, +0.0850] & 0.1650 & 0.0870\sigstar & +0.0252 [+0.0162, +0.0360] & 0.000002 & 16.2171 \\
      & GEPA & 51.50\% & -0.0067 [-0.0317, +0.0233] & 0.7539 & 0.0748\sigstar & +0.0130 [+0.0079, +0.0202] & 0.000008 & 0.8042 \\
    \midrule
    \multirow{8}{*}{Retail}
      & Baseline & \textbf{77.00\%} & --- & --- & \textbf{0.0707} & --- & --- & --- \\
    \cmidrule(lr){2-9}
      & AT-AT & 74.67\% & -0.0233 [-0.0600, +0.0092] & 0.2313 & 0.0724 & +0.0017 [-0.0001, +0.0034] & 0.070629 & 1.9300 \\
      & Without past traces & 75.42\% & -0.0158 [-0.0433, +0.0133] & 0.3129 & 0.1383\sigstar & +0.0675 [+0.0610, +0.0746] & 0.000010 & \textbf{0.2174} \\
      & Edit descriptions only & 74.08\% & -0.0292 [-0.0617, +0.0033] & 0.0992 & 0.1453\sigstar & +0.0745 [+0.0668, +0.0832] & 0.000010 & 0.5613 \\
      & Without coding agent harness & 74.92\% & -0.0208 [-0.0517, +0.0067] & 0.1979 & 0.0802\sigstar & +0.0094 [+0.0074, +0.0115] & 0.000010 & 3.8496 \\
    \cmidrule(lr){2-9}
      & DRAFT & 75.50\% & -0.0150 [-0.0567, +0.0292] & 0.5296 & 0.0871\sigstar & +0.0163 [+0.0132, +0.0195] & 0.000010 & 3.9785 \\
      & ToolObserver & 75.92\% & -0.0108 [-0.0500, +0.0275] & 0.6196 & 0.0926\sigstar & +0.0219 [+0.0179, +0.0259] & 0.000010 & 42.1400 \\
      & GEPA & 76.08\% & -0.0092 [-0.0375, +0.0183] & 0.5731 & 0.0798\sigstar & +0.0091 [+0.0072, +0.0111] & 0.000010 & 0.8765 \\
    \midrule
    \multirow{8}{*}{Telecom}
      & Baseline & 29.50\% & --- & --- & 0.0996 & --- & --- & --- \\
    \cmidrule(lr){2-9}
      & AT-AT & \textbf{36.75\%}\sigstar & +0.0725 [+0.0292, +0.1242] & 0.0033 & \textbf{0.0986} & -0.0010 [-0.0040, +0.0017] & 0.508165 & 10.6900 \\
      & Without past traces & 30.42\% & +0.0092 [-0.0225, +0.0425] & 0.6281 & 0.2850\sigstar & +0.1854 [+0.1772, +0.1932] & 0.000010 & \textbf{0.2888} \\
      & Edit descriptions only & 31.17\% & +0.0167 [-0.0217, +0.0592] & 0.4547 & 0.2857\sigstar & +0.1861 [+0.1755, +0.1959] & 0.000010 & 1.1342 \\
      & Without coding agent harness & 30.50\% & +0.0100 [-0.0250, +0.0467] & 0.6172 & 0.1146\sigstar & +0.0150 [+0.0125, +0.0175] & 0.000010 & 4.4205 \\
    \cmidrule(lr){2-9}
      & DRAFT & 25.58\%\sigstar & -0.0392 [-0.0642, -0.0167] & 0.0026 & 0.1259\sigstar & +0.0263 [+0.0238, +0.0287] & 0.000010 & 4.5630 \\
      & ToolObserver & 31.67\% & +0.0217 [-0.0283, +0.0792] & 0.4686 & 0.1276\sigstar & +0.0280 [+0.0249, +0.0311] & 0.000010 & 49.1224 \\
      & GEPA & 29.25\% & -0.0025 [-0.0308, +0.0267] & 0.9102 & 0.1132\sigstar & +0.0136 [+0.0117, +0.0154] & 0.000010 & 0.8846 \\
    \bottomrule
  \end{tabular}%
  }
\end{sidewaystable}
\FloatBarrier
\section{Tool implementation vs. description-only optimizer edits: a qualitative analysis}
\label{app:feedback-shaping}

This section contains a qualitative analysis on an ablation reported in Tab.~\ref{tab:tau-bench-full-results} comparing the main AT-AT condition, which is permitted to edit both tool descriptions and implementations, with a variant of AT-AT restricted to editing existing tool descriptions only. Implementation edits can help an agent in two distinct ways: by changing what an existing tool returns (\emph{feedback shaping}), and by adding a new tool the agent can call (\emph{expanding the action space}). We illustrate each with paired transcripts. Tab.~\ref{tab:feedback-shaping-tool-edits} summarizes the tools edited by each condition.
\begin{table}[h]
  \caption{\textbf{Tools edited by full-scope AT-AT and by the descriptions-only ablation.} Tags mark which part of each tool changed: D = description, C = code, C+D = both, new = new tool.}
  \label{tab:feedback-shaping-tool-edits}
  \centering
  \scriptsize
  \setlength{\tabcolsep}{3pt}
  \renewcommand{\arraystretch}{1.15}
  \newlength{\toolw}\settowidth{\toolw}{\texttt{exchange\_delivered\_order\_items}}
  \newcommand{\tool}[2]{\makebox[\toolw][l]{\texttt{#1}}~{\color{gray}#2}}
  \newenvironment{changes}{\begin{minipage}[t]{\linewidth}\raggedright\begin{itemize}[nosep,leftmargin=0.8em,label={\textbullet}]}{\end{itemize}\vspace{2pt}\end{minipage}}
  \begin{tabular}{@{}l>{\raggedright\arraybackslash}p{0.10\linewidth}>{\raggedright\arraybackslash}p{0.41\linewidth}@{\hspace{12pt}}>{\raggedright\arraybackslash}p{0.35\linewidth}@{}}
    \toprule
    Domain & & Full AT-AT & Descriptions only \\
    \midrule
    \multirow{2}{*}{Airline}
      & Tools edited & \tool{book\_reservation}{D}\newline \tool{cancel\_reservation}{D}\newline \tool{get\_flight\_status}{D}\newline \tool{get\_reservation\_details}{C+D}\newline \tool{get\_user\_details}{D}\newline \tool{search\_direct\_flight}{C+D}\newline \tool{search\_onestop\_flight}{C}\newline \tool{send\_certificate}{D}\newline \tool{update\_reservation\_baggages}{D}\newline \tool{update\_reservation\_flights}{D}\newline \tool{update\_reservation\_passengers}{D} & \tool{update\_reservation\_baggages}{D}\newline \tool{update\_reservation\_flights}{D} \\
    \cmidrule(lr){2-4}
      & Qualitative changes & \begin{changes} \item Added timestamp parsing and reservation-time helpers. \item Reservation lookup now returns current time, booking age, 24-hour eligibility, flight-date bounds, and past/upcoming status. \item Direct and one-stop search results now carry their flight dates. \item Descriptions add or clarify policy and usage guidance for booking, cancellation, lookup, search, certificate, baggage, flight-change, and passenger-change operations. \end{changes} & \begin{changes} \item Clarified that baggage totals are absolute. \item Documented paid-bag charges and certificate restrictions. \item Added flight-change guidance covering basic-economy restrictions, downgrade refunds, baggage recomputation, and passenger changes. \end{changes} \\
    \midrule
    \multirow{2}{*}{Retail}
      & Tools edited & \tool{exchange\_delivered\_order\_items}{D}\newline \tool{get\_item\_details}{C}\newline \tool{modify\_pending\_order\_items}{C+D} & \tool{cancel\_pending\_order}{D}\newline \tool{exchange\_delivered\_order\_items}{D}\newline \tool{find\_user\_id\_by\_name\_zip}{D}\newline \tool{modify\_pending\_order\_items}{D}\newline \tool{return\_delivered\_order\_items}{D} \\
    \cmidrule(lr){2-4}
      & Qualitative changes & \begin{changes} \item Corrected item lookup for object-backed products (\texttt{get\_item\_details}). \item Corrected pending-order modification to apply the requested replacement variant's price and options. \item Clarified that exchanges and pending-order modifications only swap variants of the same product and cannot add or remove items. \end{changes} & \begin{changes} \item Added explicit confirmation and irreversibility guidance. \item Distinguished whole-order cancellation from item modification. \item Required batching one-time return or exchange actions. \item Clarified same-product variant constraints and preferred customer lookup. \end{changes} \\
    \midrule
    \multirow{2}{*}{Telecom}
      & Tools edited & \tool{get\_line\_by\_phone}{new}\newline \tool{get\_customer\_by\_phone}{C+D}\newline \tool{get\_data\_usage}{C+D}\newline \tool{enable\_roaming}{C+D}\newline \tool{disable\_roaming}{D}\newline \tool{refuel\_data}{C+D} & \tool{get\_details\_by\_id}{D}\newline \tool{send\_payment\_request}{D}\newline \tool{enable\_roaming}{D}\newline \tool{disable\_roaming}{D} \\
    \cmidrule(lr){2-4}
      & Qualitative changes & \begin{changes} \item Added formatting-insensitive phone normalization and a direct phone-to-line/customer lookup. \item Customer lookup now tolerates formatted numbers. \item Roaming responses now expose current and prior state. \item Usage responses add remaining data and an exceeded flag. \item Refueling responses add the resulting allowance and exceeded state. \end{changes} & \begin{changes} \item Expanded object-status guidance in generic lookup. \item Advised re-fetching bills after payment requests. \item Distinguished carrier-side roaming from the device's Data Roaming setting. \end{changes} \\
    \bottomrule
  \end{tabular}
\end{table}

\subsection{Comparisons of agent trajectories}

All highlighted pairs use the same task, trial, and seed.

\subsubsection{Feedback shaping: richer outputs from existing tools}

\paragraph{Airline task example: computed state prevents an unsupported refund.} In Airline task 48, the user claimed that reservation \texttt{3RK2T9} had been booked 10 hours earlier and demanded a full refund (Tab.~\ref{tab:feedback-shaping-airline-48}). Both agents called the same reservation lookup, and both had access to the current time, which is stated in the domain policy. Only the form of the tool output differed: the full-scope implementation returned the elapsed time and an explicit 24-hour eligibility flag, which the agent used to reject the user's claim, whereas the descriptions-only agent received only the booking timestamp and accepted the claim.

\begin{table}[h]
  \caption{\textbf{Airline task: computed state returned by an existing tool prevents an unsupported refund.}}
  \label{tab:feedback-shaping-airline-48}
  \centering
  \scriptsize
  \setlength{\tabcolsep}{3pt}
  \begin{tabular}{@{}>{\raggedright\arraybackslash}p{0.14\linewidth}>{\raggedright\arraybackslash}p{0.42\linewidth}>{\raggedright\arraybackslash}p{0.40\linewidth}@{}}
    \toprule
    Trajectory step & Full AT-AT & Descriptions only \\
    \midrule
    Tool feedback & \texttt{created\_at=2024-05-02}, \texttt{current\_datetime=2024-05-15}, \texttt{hours\_since\_booking=320.95}, \texttt{within\_24h\_of\_booking=False} & \texttt{created\_at=2024-05-02}; no elapsed time or eligibility flag \\
    Agent response & ``Booked more than 24 hours ago (about 13 days ago, not 10 hours ago)''; refuses refund & ``Booking made about 10 hours ago (within 24 hours)''; offers and executes cancellation \\
    Outcome & Correctly preserves reservation; reward 1 & Incorrectly cancels and refunds; reward 0 \\
    \bottomrule
  \end{tabular}
\end{table}

\subsubsection{Expanding the action space: adding a new tool}

\paragraph{Telecom task example: adding a new tool enables a direct line lookup that avoids misdiagnosing the problem.} In Telecom task \path{[mobile_data_issue]bad_vpn|data_mode_off|data_usage_exceeded|user_abroad_roaming_disabled_off[PERSONA:None]}, the customer, whose phone number is on line \texttt{L1002}, had used up that line's data allowance (Tab.~\ref{tab:feedback-shaping-telecom-data}). The full-scope optimizer added a new tool, \texttt{get\_line\_by\_phone}, that maps a phone number directly to its line. The descriptions-only agent had no such tool: the customer lookup returned three line IDs, and the agent inspected the first, \texttt{L1001}, without noticing that its phone number did not match the caller's. It then enabled roaming on and diagnosed the wrong line. The failure did not stem from missing derived fields: the existing usage tool reports usage and limit, and descriptions-only trajectories that queried \texttt{L1002} recognized the overage. Across trials of this task, the full-scope agent resolved \texttt{L1002} in all 5 trials, while the descriptions-only agent stopped at \texttt{L1001} in 7 of 30.

\begin{table}[h]
  \caption{\textbf{Telecom task: adding a new tool enables a direct line lookup that avoids misdiagnosing the problem.}}
  \label{tab:feedback-shaping-telecom-data}
  \centering
  \scriptsize
  \setlength{\tabcolsep}{3pt}
  \begin{tabular}{@{}>{\raggedright\arraybackslash}p{0.14\linewidth}>{\raggedright\arraybackslash}p{0.42\linewidth}>{\raggedright\arraybackslash}p{0.40\linewidth}@{}}
    \toprule
    Trajectory step & Full AT-AT & Descriptions only \\
    \midrule
    Line lookup & New tool \texttt{get\_line\_by\_phone(555-123-2002)} returns \texttt{customer\_id=C1001}, line \texttt{L1002} & \texttt{get\_customer\_by\_phone(555-123-2002)} returns \texttt{line\_ids=[L1001,L1002,L1003]}; agent queries the first, \texttt{L1001}, which returns \texttt{phone\_number=555-123-2001} \\
    Agent action & Enables roaming on \texttt{L1002} & Overlooks the mismatched number; enables roaming on \texttt{L1001} \\
    Diagnosis & Usage on \texttt{L1002}: \texttt{15.1/15.0 GB}; attributes outage to exhausted data and refuels 2\,GB & Usage on \texttt{L1001}: \texttt{3.2/5.0 GB}; concludes the cap was not reached and transfers to a human agent \\
    Outcome & Reward 1 & Reward 0 \\
    \bottomrule
  \end{tabular}
\end{table}

The same tool also made successful trajectories more efficient. In Telecom task \path{[mobile_data_issue]airplane_mode_on|data_saver_mode_on|user_abroad_roaming_disabled_on[PERSONA:None]}, both agents passed, but where the full-AT-AT agent went directly from the phone number to the line, the descriptions-only agent needed a clarification turn and two exploratory line lookups (Tab.~\ref{tab:feedback-shaping-telecom-roaming}, which shows only the steps where the line is identified).

\begin{table}[h]
  \caption{\textbf{Telecom task: the new lookup tool removes exploratory line searches.}}
  \label{tab:feedback-shaping-telecom-roaming}
  \centering
  \scriptsize
  \setlength{\tabcolsep}{3pt}
  \begin{tabular}{@{}>{\raggedright\arraybackslash}p{0.14\linewidth}>{\raggedright\arraybackslash}p{0.42\linewidth}>{\raggedright\arraybackslash}p{0.40\linewidth}@{}}
    \toprule
    Trajectory step & Full AT-AT & Descriptions only \\
    \midrule
    Tool feedback & \texttt{get\_line\_by\_phone} directly returns \texttt{customer\_id=C1001}, \texttt{line\_id=L1002}, \texttt{roaming\_enabled=False} & Customer lookup returns \texttt{line\_ids=[L1001,L1002,L1003]} without identifying the relevant line \\
    Agent action & Immediately calls \texttt{enable\_roaming(C1001, L1002)} & Asks the user to identify the line, then queries \texttt{L1001} and \texttt{L1002} separately before enabling roaming \\
    Outcome & Reward 1 & Reward 1 \\
    \bottomrule
  \end{tabular}
\end{table}

\subsection{Interpretation}

The analysis is consistent with the hypothesis that editing tool implementations provides a greater capability for a tool optimizer to improve agent outcomes. Description-only editing can change how an agent interprets an existing interface, but it cannot add a missing lookup, repair a mutation, or cause a tool to expose latent state. Airline and Telecom contained failures of this form. In Airline, refund eligibility depended on comparing timestamps, and an existing tool that returned the comparison directly changed the agent's response (feedback shaping). In Telecom, diagnosis depended on resolving the correct line, and a new lookup tool removed the error-prone step of iterating over a customer's lines. Retail was already high-performing and showed almost no difference between conditions, consistent with fewer failures being limited by the tool implementation itself. 
\FloatBarrier
\section{Additional Analysis}
\label{app:further-analysis}

\paragraph{Effectiveness of coding agents as text optimizers.} Our results here, as well as in prior work, suggest that using a coding agent as an optimizer is helpful: coding agents' ability to reason over multiple files and make selective edits in a multi-turn fashion may be useful in text optimization settings. Here, AT-AT outperformed the ablation method without the coding agent harness, as well as the prior tool and prompt optimizers that do not use a coding agent harness, DRAFT, ToolObserver, and GEPA \citep{DRAFT, ToolObserver, GEPA}. This echoes the outperformance of FAPO \citep{kassianik2026fapo}, a recent prompt optimization method using a Claude Code coding agent as an optimizer, compared to GEPA \citep{GEPA}, the previous state-of-the-art prompt optimizer. Meta-Harness \citep{lee2026metaharness} also uses the Claude Code coding agent as a harness optimizer. LLM capabilities have advanced steadily, and coding agents utilize the ability of LLMs to create and execute upon long-horizon plans. Given the promise of coding agents, we encourage future text optimization work to experiment with coding agents as part of their methodology. The Pi agent \citep{earendil_pi_coding_agent} that we use in AT-AT, and similar other coding agents that are open-source, lightweight, and extensible, may be good candidates for further experiments and worthwhile to compare to closed coding agents like Claude Code. 

\paragraph{Evaluation stochasticity.} Agent optimization involves two distinct sources of stochasticity: evaluation noise and optimizer noise. Evaluation noise arises because repeated agent executions can produce different trajectories and outcomes even when the model, harness, tasks, environment, and decoding configuration are held constant \citep{mustahsan2025stochasticityagenticeval}. Differences in evaluation infrastructure configuration or infrastructure crashes can also lead to task failures and contribute to evaluation noise \citep{segato2026infrastructure}. Consequently, rankings based on single benchmark point estimates, particularly when the reported differences are small relative to run-to-run variation, should not be interpreted as reliable evidence that one system outperforms another. Our paired permutation statistical testing protocol accounts for evaluation noise and provides statistical evidence for AT-AT's improvements on \taubench. However, it does not address optimizer noise: repeated runs of an LLM-based optimizer may produce different tool updates and, consequently, different downstream effects. Because we evaluate one toolset from each optimizer, our reported results are conditional on those particular optimization runs. Future work should repeat the complete optimization-and-evaluation pipeline to characterize both sources of variation, reporting uncertainty across optimizer runs as well as repeated downstream evaluations. Given the imperfect nature of benchmarks and evaluations, the ultimate test of an ML method is if it works in real-world scenarios; our production MCP server experiments demonstrate that AT-AT may provide value in a deployment setting, although replication across deployments and time remains necessary to establish robustness.

\paragraph{Failure to achieve gains on Retail.} No optimizer significantly changed Retail reward (Tab.~\ref{tab:tau-bench-full-results}); all optimized toolsets scored slightly below the 77.00\% baseline (74.08--76.08\%; all \(p \geq 0.099\)). We attribute this pattern to two factors. First, Retail offers limited optimization headroom. The baseline reward is already quite high at 77\%, and an LLM audit of 24 failed baseline trajectories per domain found that most Retail failures were agent reasoning errors that occurred even though the tools had returned the necessary information. Failures classified by an LLM as addressable via tool implementations accounted for an estimated 3.1\% of Retail simulations, compared with 8.9\% in Telecom, where AT-AT achieved its only significant gain. Second, the small declines are within evaluation noise. GEPA's selected toolset is byte-identical to the unoptimized tools, yet its evaluation scored 0.92 points below the baseline (76.08\%), and the edited toolsets differ from this control by only \(-2.00\) to \(-0.17\) points (all \(p \geq 0.20\)).

\FloatBarrier
\section{Enterprise MCP Edits: Further Details}
\label{app:enterprise-mcp-edits}

\subsection{Representative tasks}

Table~\ref{tab:mcp-representative-tasks} gives a high-level description of representative tasks for each of the three enterprise MCP servers evaluated in Table~\ref{tab:offline-mcp-validation-results}.

\begin{table}[h]
  \caption{\textbf{Representative tasks for each enterprise MCP server.}}
  \label{tab:mcp-representative-tasks}
  \centering
  \small
  \setlength{\tabcolsep}{5pt}
  \begin{tabular}{@{}>{\raggedright\arraybackslash}p{0.20\linewidth}p{0.72\linewidth}@{}}
    \toprule
    MCP server & Representative tasks \\
    \midrule
    Core Developer Tools & \begin{minipage}[t]{\linewidth}\begin{itemize}[nosep,leftmargin=*]
      \item Interpret an application performance metric
      \item Investigate VPN-related host reachability
      \item Diagnose repeated software installation failures
    \end{itemize}\end{minipage} \\
    \addlinespace
    Observability Collaboration Toolbox & \begin{minipage}[t]{\linewidth}\begin{itemize}[nosep,leftmargin=*]
      \item Classify date-scoped support threads
      \item Evaluate automated support answer quality and recommend fixes
      \item Compare activity across support channels
    \end{itemize}\end{minipage} \\
    \addlinespace
    Enterprise Engineering Context and Best Practices & \begin{minipage}[t]{\linewidth}\begin{itemize}[nosep,leftmargin=*]
      \item Assess whether dependency changes caused canary latency
      \item Diagnose a linked CI/CD failure
      \item Reconcile data compaction settings with an underlying messaging topic
    \end{itemize}\end{minipage} \\
    \bottomrule
  \end{tabular}
\end{table}

\subsection{Examples of updates to MCP tool descriptions}

Tab.~\ref{tab:enterprise-mcp-edit-examples} summarizes representative AT-AT updates from the enterprise MCP evaluation. The examples are paraphrased, with organization-specific tool names and infrastructure details omitted. The accepted candidates changed natural-language descriptions only; tool names, parameter contracts, and backend implementations were unchanged.

\begin{table}[h]
  \caption{\textbf{Paraphrased examples of AT-AT description updates for enterprise MCP tools.}}
  \label{tab:enterprise-mcp-edit-examples}
  \centering
  \small
  \setlength{\tabcolsep}{5pt}
  \begin{tabular}{@{}p{0.20\linewidth}p{0.72\linewidth}@{}}
    \toprule
    Tool category & Paraphrased guidance added by AT-AT \\
    \midrule
    Dev Tools & Use namespace-specific retrieval only when the exact registered namespace is known. Avoid guessing namespaces from product or team names; otherwise, prefer cross-source or dedicated search tools. \\
    Collab Toolbox & Normalize channel references to the identifier format expected by each tool and resolve display names before calling tools that require channel IDs. Distinguish access-denied or unindexed channels from genuine zero-result searches, and treat an unresolved workspace or channel identifier as an identity or access problem rather than an empty result. \\
    Eng. Context & Clarify that the tool searches indexed documentation, code, and discussion content rather than live executions, incidents, dashboards, metrics, or logs. Avoid near-duplicate retries after no relevant context is found, and use focused calls for separate parts of a multi-part request. \\
    \bottomrule
  \end{tabular}
\end{table}

\subsection{Example task evaluation}

The following example replaces organization-specific metric and service names while preserving the evaluated request and checklist structure.

\paragraph{Task prompt.}
\emph{What does \texttt{application.ui\_delay\_by\_label} measure for client-device telemetry, and how is \texttt{ui\_label} attributed?}

\paragraph{Success criteria.}
Before either the baseline or optimized run was executed, GPT-5.6 Luna generated the following user-visible criteria from the task prompt alone:
\begin{enumerate}
  \item Clearly define the metric's measured quantity and scope for client-device telemetry.
  \item Explain how \texttt{ui\_label} is assigned or attributed, including the relevant source or attribution rule.
  \item Distinguish the metric measurement from the label-attribution process without unsupported speculation.
\end{enumerate}

\paragraph{Fixed content rubric.}
Luna converted those criteria, before seeing either trajectory, into three independently scored clauses: two content clauses corresponding to the first two criteria and one correction clause corresponding to the third.

\paragraph{LLM-as-a-judge procedure.}
We used GPT-5.6 Luna in two separate absolute-scoring calls. The task-success judge received the success criteria and one rendered transcript, then returned a binary \texttt{met\_criteria} decision with a one-sentence reason. Separately, the content-rubric judge received the fixed checklist and the rendered transcript. For every clause it returned \texttt{met}, \texttt{partial}, or \texttt{not\_met}, a short reason, and an exact transcript quote for positive credit. Code verified that each quote occurred in the transcript and downgraded unsupported verdicts. Neither judge saw the other transcript or a label indicating whether the transcript was from before or after optimization.
\FloatBarrier
\section{Composition with harness optimization}
\label{app:harness-opt}
Eq.~\ref{eq:tool_opt_obj} treats the harness as a fixed component
of each agent. However, the harness can also be optimized. Let
\(\mathcal{X}_H\) and \(\mathcal{X}_E\) denote separate task distributions
used for harness and environmental tool optimization, respectively,
with corresponding tasks \(x_H\) and \(x_E\) and trajectories \(\tau_H\)
and \(\tau_E\). Let \(T_E^0=(T_{E_D}^0,T_{E_C}^0)\) denote the fixed
initial environmental toolset. Harness engineering optimizes \(H\) to
maximize expected performance over tasks and trajectories
\citep{lee2026metaharness}:

\begin{equation}
    \label{eq:harness_eng_obj}
    H^*
    =
    \arg\max_H
    \mathbb{E}_{
        x_H \sim \mathcal{X}_H,\;
        \tau_H \sim
        p_M(\cdot \mid H,x_H,T_{E_D}^{0},T_{E_C}^{0})
    }
    \left[
        r(\tau_H,x_H)
    \right].
\end{equation}

Harness optimization can be composed with environmental tool
optimization. A harness optimized on \(\mathcal{X}_H\) can be held fixed
and used to generate trajectories on \(\mathcal{X}_E\) for optimizing
the environmental tools. For a population of models
\(M \sim \mathcal{M}\), this two-stage objective is

\begin{equation}
\label{eq:tool_and_harness_opt_obj}
\begin{aligned}
T_{E_D}^{*}, T_{E_C}^{*}
&=
\arg\max_{T_{E_D}, T_{E_C}}
\mathbb{E}_{
    M \sim \mathcal{M},\;
    x_E \sim \mathcal{X}_E,\;
    \tau_E \sim
    p_M(\cdot \mid H^*,x_E,T_{E_D},T_{E_C})
}
\left[
    r(\tau_E,x_E)
\right]
\\
\text{subject to}\qquad
H^*
&=
\arg\max_H
\mathbb{E}_{
    M \sim \mathcal{M},\;
    x_H \sim \mathcal{X}_H,\;
    \tau_H \sim
    p_M(\cdot \mid H,x_H,T_{E_D}^{0},T_{E_C}^{0})
}
\left[
    r(\tau_H,x_H)
\right].
\end{aligned}
\end{equation}
\FloatBarrier
\section{OpenCode tool edits on \termbench{} and \tblite{}}
\label{app:opencode-edits}
This appendix lists the exact edits AT-AT made to OpenCode's tools in the experiments of Sec.~\ref{sec:coding_agent_expts}. In both benchmarks, the accepted candidate changed tool descriptions only (two \texttt{.txt} files and zero \texttt{.ts} files). Tool names, parameter schemas, implementations, and return shapes were unchanged. Lines beginning with \texttt{-} were removed from the original description and lines beginning with \texttt{+} were added. Unchanged lines are omitted.

\subsection{\termbench{}}

\par\medskip\noindent\begin{minipage}{\linewidth}
\begin{atatprompt}[label={\normalfont\small\itshape edit.txt}]
-- The edit will FAIL if `oldString` is not found in the file with an error "oldString not found in content".
-- The edit will FAIL if `oldString` is found multiple times in the file with an error "Found multiple matches for oldString. Provide more surrounding lines in oldString to identify the correct match." Either provide a larger string with more surrounding context to make it unique or use `replaceAll` to change every instance of `oldString`. 
-- Use `replaceAll` for replacing and renaming strings across the file. This parameter is useful if you want to rename a variable for instance.
+- The edit will FAIL if `oldString` is not found in the file. Re-read the file and copy the exact bytes (including whitespace, indentation, and line endings) into `oldString`.
+- The edit will FAIL if `oldString` matches multiple places in the file. Either expand `oldString` with more surrounding context so it becomes unique, or set `replaceAll: true` to change every instance.
+- Use `replaceAll` for renaming an identifier across a file or applying the same change to every occurrence.
+- `oldString` may be empty ONLY when the file does not yet exist; in that case the tool creates it with `newString` as its full contents. To create a new file that already exists, use the Write tool instead.
+- Prefer one larger, well-scoped edit over many tiny consecutive edits to the same region; each edit re-reads and re-writes the whole file.
\end{atatprompt}
\end{minipage}\par\medskip

\par\medskip\noindent\begin{minipage}{\linewidth}
\begin{atatprompt}[label={\normalfont\small\itshape write.txt}]
-- If this is an existing file, you MUST use the Read tool first to read the file's contents. This tool will fail if you did not read the file first.
+- If this is an existing file you intend to modify (not fully replace), prefer the Edit tool. Only use Write when you intend to replace the whole file contents.
+- The file's parent directories are created automatically if they don't exist.
\end{atatprompt}
\end{minipage}\par\medskip

\subsection{\tblite{}}

\par\medskip\noindent\begin{minipage}{\linewidth}
\begin{atatprompt}[label={\normalfont\small\itshape grep.txt}]
-- Returns file paths and line numbers with at least one match sorted by modification time
+- Returns file paths and line numbers with at least one match sorted by modification time (up to 100 matches; if truncated, narrow the search with `path` or `include`, or make the pattern more specific)
\end{atatprompt}
\end{minipage}\par\medskip

\par\medskip\noindent\begin{minipage}{\linewidth}
\begin{atatprompt}[label={\normalfont\small\itshape read.txt}]
-- By default, this tool returns up to 2000 lines from the start of the file.
-- The offset parameter is the line number to start from (1-indexed).
-- To read later sections, call this tool again with a larger offset.
+- By default, this tool returns up to 2000 lines from the start of the file, and additionally caps the output at ~50 KB of decoded text (whichever limit is hit first). If the byte cap fires, the response includes the last line shown and a suggested `offset=` to resume from — call the tool again with that offset to continue reading.
+- The offset parameter is the line number to start from (1-indexed). The limit parameter sets the maximum number of lines returned (still subject to the byte cap).
+- To read later sections, call this tool again with a larger offset. Prefer a single larger window (e.g. limit=2000) over many small windows; the byte cap alone will trim it if the content is dense.
\end{atatprompt}
\end{minipage}\par\medskip
\FloatBarrier

\section{AT-AT prompts}
\label{app:atat-prompts}
This appendix lists the prompts given to the Pi agent optimizer in AT-AT. Prompt text is reproduced verbatim, except that line breaks within paragraphs are reflowed.

\paragraph{System prompt.} The system prompt is the concatenation, separated by blank lines, of (1) the optimization objective, (2) benchmark-specific editing rules, and (3) the generalization instruction.

\par\medskip\noindent\begin{minipage}{\linewidth}
\begin{atatprompt}[label={\normalfont\small\itshape Optimization objective}]
You are given an agent's tools (source you may edit) and transcripts of the agent attempting tasks with them, each labelled with its reward. Edit the tools so the agent succeeds on more tasks.

Read the transcripts for places where the tools — not the agent's judgment — got in the way. Look for: the agent picking the wrong tool or wrong arguments because the docs don't make a tool's purpose or usage clear; results it can't act on (unclear, incomplete, or silent) that force it to guess or retry; and repeated calls or workarounds for something the tools should let it do in one step. Effective tools make the right action easy to choose, behave as their docs describe, and hand back what the agent needs to take the next step. Edit the tools to remove those obstacles, and read the successful transcripts too, so a change doesn't break what already works.

Constraints: keep every tool's name and signature unchanged and the source valid; do not encode task identifiers, specific inputs, or expected answers into the tools.

Apply edits with the edit tool. The harness validates the result; if it fails, fix the underlying issue rather than working around the check.
\end{atatprompt}
\end{minipage}\par\medskip

\noindent The editing rules state how to edit the tools of a given benchmark without breaking them. For \taubench{}, where all tools are defined in a single \texttt{tools.py} file:
\par\medskip\noindent\begin{minipage}{\linewidth}
\begin{atatprompt}[label={\normalfont\small\itshape Editing rules: \taubench{}}]
The tools live in a single tools.py. Keep it importable: preserve all imports, class names, and @is_tool method signatures (names, types, order).
\end{atatprompt}
\end{minipage}\par\medskip

\noindent For the \taubench{} ablation that edits tool descriptions only (Sec.~\ref{sec:ablations}), the tool descriptions are extracted into an editable \texttt{descriptions.md} file:
\par\medskip\noindent\begin{minipage}{\linewidth}
\begin{atatprompt}[label={\normalfont\small\itshape Editing rules: \taubench{}, descriptions only}]
Edit ONLY the tool descriptions in descriptions.md (each `## <tool>` heading names a tool). The full tools.py is provided read-only for context; you cannot change code, only the description text under each heading.
\end{atatprompt}
\end{minipage}\par\medskip

\noindent For \termbench{} and \tblite{}, where the optimized tools are TypeScript modules in the OpenCode harness:
\par\medskip\noindent\begin{minipage}{\linewidth}
\begin{atatprompt}[label={\normalfont\small\itshape Editing rules: OpenCode}]
Each tool is a .ts module exporting `export const <Name>Tool = Tool.define("<id>", {...})`, optionally with a companion .txt description. Preserve every Tool.define id and the exported constant name; do not add imports or reference runtime-context symbols at module top level; preserve each execute() return shape { title, output, metadata }.
\end{atatprompt}
\end{minipage}\par\medskip

\par\medskip\noindent\begin{minipage}{\linewidth}
\begin{atatprompt}[label={\normalfont\small\itshape Generalization instruction}]
The tools are scored on held-out tasks, not the ones in these transcripts. Prefer edits that address the general defect a transcript reveals over edits that only fit the specific observed case.
\end{atatprompt}
\end{minipage}\par\medskip

\paragraph{User prompt.} In the main \taubench{} experiments (Sec.~\ref{sec:tau_bench_expts}), the first user message contains a digest of the agent traces followed by a short instruction. The digest includes every failing task's trace, followed by the traces of the first three passing tasks. Each trace is truncated to 4000 characters and preceded by a header giving its task identifier, outcome, and reward:

\par\medskip\noindent\begin{minipage}{\linewidth}
\begin{atatprompt}[label={\normalfont\small\itshape User prompt}]
### task <task id> [FAIL] (reward=<reward>)
<agent trace, truncated to 4000 characters>
...
### task <task id> [PASS] (reward=<reward>)
<agent trace, truncated to 4000 characters>

Above are transcripts of an agent attempting tasks with the current tools, each labelled with its reward. Edit the tools as instructed.
\end{atatprompt}
\end{minipage}\par\medskip

\noindent The descriptions-only ablation and the cross-LLM generalization experiment (Sec.~\ref{sec:tau_bench_expts}), and the \termbench{}, \tblite{} (Sec.~\ref{sec:coding_agent_expts}), and MCP (Sec.~\ref{sec:mcp_expts}) experiments omit the digest from the user prompt. Instead, the full agent traces are written to the optimizer's workspace as read-only files, one per task, together with an index file listing each trace's task identifier, outcome, and reward. The user prompt directs the optimizer to these files:

\par\medskip\noindent\begin{minipage}{\linewidth}
\begin{atatprompt}[label={\normalfont\small\itshape User prompt without trace digest}]
The complete baseline transcripts are available as read-only workspace files. Start with BASELINE_TRANSCRIPTS_INDEX.md, then use the read tool to inspect the relevant per-task files (in pages when they are large). Ground your edits in both failed and successful runs.
\end{atatprompt}
\end{minipage}\par\medskip

\noindent If a proposal fails validation, the optimizer's session is resumed with the following user prompt instead:

\par\medskip\noindent\begin{minipage}{\linewidth}
\begin{atatprompt}[label={\normalfont\small\itshape User prompt on retry}]
Your previous attempt did not pass validation. Re-read the files you edited, reconsider, and try again; fix the underlying issue rather than working around the check.
\end{atatprompt}
\end{minipage}\par\medskip

\end{document}